\RequirePackage{amsmath}
\documentclass[final, runningheads, 11pt]{llncs}

\usepackage{comment}
\usepackage{amssymb}
\usepackage{tabularx}

\PassOptionsToPackage{hyphens}{url}
\usepackage[hidelinks]{hyperref}
\usepackage[T1]{fontenc}
\usepackage{graphicx}
\usepackage{color}

\usepackage{amsmath}
\newcommand{\itadata}{\footnotesize \textsl{ITADATA2024: The 3$^{\text{rd}}$ Italian Conference on Big Data and Data Science}}
\usepackage{fancyhdr}
\fancypagestyle{empty}{\fancyhf{}\fancyhead[C]{\itadata}
  \fancyhead[R]{\includegraphics[width=.05\textwidth]{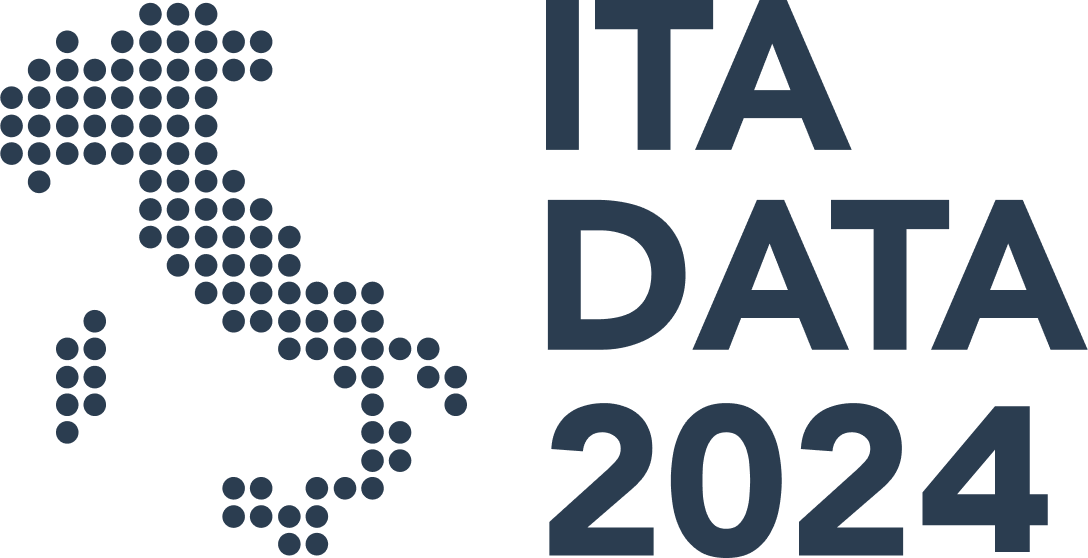}}}

\makeatletter
\begin{document}
\setlength{\headheight}{13.59999pt}

\title{Active Data Sampling and Generation for Bias Remediation}
\author{Antonio Maratea\orcidID{0000-0001-7997-0613} \and
Rita Perna
}
\authorrunning{A. Maratea and R. Perna}
\institute{University of Naples Parthenope, Centro Direzionale Isola C4, 80147 Napoli, IT 
\email{antonio.maratea@uniparthenope.it}\\
\email{rita.perna001@studenti.uniparthenope.it}\\
\url{https://scienzeetecnologie.uniparthenope.it/} }
\maketitle

\begin{abstract}
Adequate sampling space coverage is the keystone to effectively train trustworthy Machine Learning models. 
Unfortunately, real data do carry several inherent risks due to the many potential biases they exhibit when gathered without a proper random sampling over the reference population, and most of the times this is way too expensive or time consuming to be a viable option. 
Depending on how training data have been gathered,  unmitigated biases can lead to harmful or discriminatory consequences that ultimately hinders large scale applicability of pre-trained models and  undermine their truthfulness or fairness expectations.
In this paper, a mixed active sampling and data generation strategy --- called \emph{samplation} --- is proposed
as a mean to compensate during fine-tuning of a pre-trained classifer the unfair classifications it produces, assuming that the training data come from a non-probabilistic sampling schema.
Given a pre-trained classifier, first a fairness metric is evaluated on a test set, then new reservoirs of labeled data are generated and finally a number of  reversely-biased artificial samples are generated for the fine-tuning of the model. 
Using as case study Deep Models for visual semantic role labeling, the proposed method has been able to fully cure a simulated gender bias starting from a 90/10 imbalance, with only a small percentage of new data and with a minor effect on accuracy.

\end{abstract}
\section{Bias in AI}
The social impact of Machine Learning algorithms nowadays requires in many cases to address biases and enforce \textbf{fairness}, meaning that they must not discriminate against particular individuals or groups  on the basis of various sensitive attributes protected by the law.
This definition of bias has remote roots that go back to the early days when the first algorithms were developed \cite{bias}. 

However, the word ``bias" has at least six partially overlapping meanings, often used interchangeably, depending on the context:
\begin{enumerate}
    \item a first technical meaning, that is the difference between the true value of a statistics on a population and the estimated value of the same statistics obtained through an estimator;
    \item another technical meaning, that stands for ``systematic error" in a measurement, due for example to the different calibration of instruments or to external sources of variability embedded in the measures;     
    \item a synonym for ``prejudice";
    \item a synonym for ``imbalance";
    \item a synonym for ``unfair", where a decision process results in unfair outcomes against certain individuals or groups;
    \item a weight used for sampling units, that gives more relevance to some units (as in \emph{biased sampling}).
\end{enumerate}
These meanings may well contradict each other, that is an unbiased estimator in technical sense may well be unfair or a fair estimator may result from biased sampling, depending on the reference population composition.
In a broad sense, technical bias represents distance from the true value, where the focus is on the accuracy of the model,  whereas prejudices referred to as "biases" represent distance from an ideal, arbitrary value, established by an authority and \emph{a priori} given, where the focus is on the fairness of the model, according to a desired target value.

In many cases, the presence of bias can be traced back to the \emph{data generation} and \emph{data collection} phases. Strictly concerning training data, one unavoidable source of bias is due to the sampling process by itself: 
\emph{selection bias} is a technical bias that arises from the selection of particular instances rather than others and, consequently, from a  dataset that is under-representative for a fraction of the population, resulting in a poor generalization of the learned algorithm \cite{bias}. 

Moreover, selection bias can generate \emph{underspecified} datasets, where multiple plausible hypotheses can describe the data equally well due to lack of relevant information to distinguish them. This can by itself lead to \emph{spurious correlations} in the data, i.e., relationships that seem significant but are actually casual --- neither due to correlation nor to causality --- which can be captured and amplified by a wrong model \cite{ye2024spurious}. 

Selection bias is very common and has many subtypes.  
For example, filling in an online questionnaire on the computer use is much more likely to attract people interested in technology than the rest of the population \cite{hellstrom2020bias}, and this is called \textbf{self-selection bias}. 
Budget constraints and ready to use benchmark datasets bind researchers to available data, that very often suffer from selection biases. 

Moreover, biases absorbed through the data generation and collection phases are propagated in the Machine learning pipeline and amplified in its outcomes \cite{ntoutsi2020bias}, yielding a self-reinforcing feedback loop, so it is paramount to address biases as early as possible.

\section{Related work}
Bias mitigation methods that can be applied at specific points of the ML pipeline: before, during, or after the model training. 

\begin{itemize}
    \item \textbf{pre-processing} methods: act at the data level and start from the consideration that data itself are biased, in the sense that the distribution over the protected attribute or any other variable is skewed or imbalanced \cite{caton2024fairness}. To address the data bias, these methods alter the original distribution of the dataset by applying particular transformations in an attempt to make it representative of the entire population. 
    Typical pre-processing techniques involve using \textbf{generative adversarial networks} (GANs) or whatever \textbf{data augmentation} techniques to create synthetic samples for balancing the training data; adopting sampling strategies to ``break down'' spurious correlations by \textbf{oversampling} or \textbf{undersampling} particular data through repetition or deletion of specific samples, respectively \cite{qraitem2023bias}\cite{zhao2023men}, and make them more representative to be learned from a model. 
    The key of sampling approaches is to decouple the training data into subgroups (or combinations), established by a-priori analysis via in- or post-processing methods, of one or more variables (e.g., \{white, female\}) and try to make them equally balanced \cite{caton2024fairness}. 
    Undersampling drops samples from specific subgroups, which involve samples sharing the same attribute $o$ and the same group $g$, to balance the overall distribution of the dataset. Assuming to have a dataset $D$, an attribute $o \in O$ and a group $g \in G$, this sampling strategy acts by eliminating samples from each subgroup such that its size is reduced to that of the subgroup with the smallest size, $\underset{c,s}{\min} \left\vert D_{c,s} \right\vert$. 
    On the contrary, oversampling acts by repeating copies of samples in specific subgroups such that their size match that of the subgroup with the highest size. 
    Another option, in addition to the previous two, could be the \textbf{upweighting} which assigns higher weights to specific samples (e.g., samples from the minority classes) in order to give them stronger contribution during the training phase \cite{qraitem2023bias}. 
    After that, the model would therefore be trained on this resulting``repaired'' dataset \cite{caton2024fairness}. \\
    Other pre-processing transformations involve \textbf{relabelling} and \textbf{perturbation}. Relabelling consists in modifying the labels of training data (and sometimes test data) in a way that the proportion of positive instances is equal across all the protected groups. 
    Perturbation, instead of operating on the labels, creates variations of the input data itself \cite{caton2024fairness}.
    
    \textbf{pre-processing} methods: act at the data level, thus before they are given as training data to a model. This class of methods start from the observation that the distribution over the protected attribute or any other variable is skewed or imbalanced. 
    To address the data bias, these methods alter the original distribution of the dataset by applying some transformations in an attempt to make it representative of the entire population.
    Typical pre-processing techniques involve using generative adversarial networks (GANs) or other data augmentation techniques to create synthetic samples for balancing the training data or 
    adopting sampling strategies to “break down” spurious correlations by 
    oversampling or undersampling particular data through repetition or deletion of specific samples, respectively.
    Another option is the upweighting, which assigns higher weights to specific samples (e.g., samples from the minority classes) in order to give them stronger contribution during the training phase. 
    Other pre-processing transformations involve relabelling and perturbation. Relabelling consists in modifying the labels of training data (and sometimes test data) in a way that the proportion of positive instances is equal across all the protected groups. Perturbation, instead of operating on the labels, creates variations of the input data itself.
    
    \item \textbf{in-processing} methods: aim to change the learning procedure of the model by incorporating one or more fairness metrics into the optimization functions or imposing a constraint, in a bid to converge toward a model parameterization that maximizes performance and fairness.
    Classically, regularization penalizes the complexity of the ML model to inhibit overfitting, but it could be also extended by adding a penalty/fairness term in the loss function which penalizes unfair outcomes \cite{caton2024fairness}.
    As pre-processing, another option is represented by \textbf{adversarial learning} whose aim for an adversary is to determine whether the training process is sufficiently fair, and if not, feedback from the adversary is used to improve the model \cite{caton2024fairness}. The fairness notion is included in the adversary to apply feedback for model tuning, and this is achieved by formalizing a multi-constraint optimization problem. 
    Model-based solutions (or non-sampling solutions) commonly suggested in literature include \textbf{corpus-level constraints} to ensure that inference predictions follow a desired distribution \cite{wang2020fairness}; \textbf{adversarial debiasing} that aligns with the concept of \emph{fairness through blindness}, so that if a variable is suspected to bias the model, the model should not consider it. 
    Here, the model is trained according to \emph{minimax} objective where the classifier aims to maximize its ability to predict the target label, while the adversary aims to minimize its ability to predict the protected variable based on the learned features \cite{wang2020fairness}. This adversarial setup encourages the model to learn features that are predictive of the target variable while being uninformative with respect to the protected variable; and \textbf{domain-independent training} which promotes \emph{fairness through awareness} and attempts to learn to differentiate between the same attribute for different groups \cite{zhao2023men}.
    A rich literature on so-called fairness-aware Machine Learning methods has flourished (see \cite{faml} for a survey).

    \item \textbf{post-processing} methods: treat the model as a black-box and apply transformations to the model predictions to improve fairness, generally by reassigning the labels based on a certain function \cite{mehrabi2021survey} i.e., the so-called relabelling.
    Other post-processing methods involve calibration, which aims to adjust the probability outputs of a model such that the proportion of positive predictions is equal to the proportion of positive samples in the dataset, and thresholding which consists in determining appropriate threshold values for each subgroup to find a balance between the true and false positive rates \cite{caton2024fairness}. Threshold values are simply those values above or below which a model makes a positive or negative prediction.
\end{itemize}

It is difficult to determine which type of approach is suitable for a given scenario, but in general it can be observed that all the model-based (i.e., in-processing) solutions require a greater effort in terms of architectural changes to the model itself or the inclusion of additional loss functions with more hyper-parameter tuning, while data sampling techniques from the probabilistic sampling or class imbalance literature aim to solve the problem at the source and are application-independent \cite{qraitem2023bias}. 
Pre-processing and post-processing are more immediate and easier solutions, rather than in-processing ones, as they do not require a study and comprehension of the underlying neural model in order to make changes to it. However, they also present limitations and challenges: pre-processing data can be time-consuming and may not always be effective, especially if the data used to train models is already biased, while post-processing decisions is a very complex task and requires a large amount of additional data \cite{ferrara2023fairness}.

It must also be stressed that bias mitigation methods also present legal implications, in that modifying the training data and/or model results according to a targeted value may violate the law and affect the model interpretability (see \cite{caton2024fairness}).

\subsection{Active Data Sampling for Bias Remediation}

Optimality of the sample is a somewhat controversial definition, and in the traditional sense it is interwoven with sample representativeness and \emph{probabilistic sampling design}.
When swapping into the domain of non-probabilistic sampling, the data  exhibit a  \emph{selection bias} from the beginning, due to the uncontrolled way in which they have been gathered. 
This is always the case when working with the data at hand without a precisely defined reference population, or when the time or budget constraints force to have quick answers on the base of the data that are easier to get.

While the disproportion among some groups may well be present in the population and hence perfectly reproduced in the samples (that ultimately must adhere to the reality to be trustable), it can also results as an artifact of the data gathering process in itself, if the sample is not assembled following a proper probabilistic sampling design. 
In classification, under-representation of some groups or over-representation of others are well known issue that can cause imbalanced learning and unfair outcomes.
It must be stressed again that from a descriptive and technical perspective the imbalance is not necessarily an issue, if it reflects the composition of the reference population, while from a prescriptive and fair perspective it is, if it does not comply to an \emph{a priori} given standard. The discussion of the legitimate criteria for the definition of this standard is beyond the scope of this paper.

\subsection{Sampling}
Given a population $\mathcal{P}$ of potentially infinite statistical units $u_j\in \mathcal{P}, \forall j\in{\{1,\ldots,M\}}$, for each unit $u_j$ a vector of features $f_{k,j}\in F_k, \forall k\in{\{1,\ldots,O\}}$ may be measured and some derived \emph{statistics} $\Theta(f_{k,j}), \forall k,j$  may be computed on the whole population data. 
Statistical units are the instances and $O$ represents the number of measured variables on each instance.
For example $\mathcal{P}$ may be the scientists currently working on Deep Learning, $F_1$ may be their current $H$-index, $F_2$ the number of publications they made in the current year and  $\Theta(f_{k,j}), \forall k,j$ may be the average Impact Factor IF of the whole Deep Learning community --- just one number for the whole population. The only way to obtain the true value of this number is to fully scan the population data, an unfeasible option most of the times due to multiple reasons: the population inclusion criteria may be ambiguous or evolving (as in the example above); the population size may be too big or unknown; many data can be missing; many units may be unreachable; there are anyway cost or time constraints to be accounted for; the disclosure of data requires a consent; the study is based on voluntary participation; etc.

Probabilistic sampling theory is one century old and provides a plethora of well-grounded, efficient and effective methods to estimate global statistics of a population starting from a finite subset of its units, called a \emph{sample}, with a reasonable and predefined confidence level, when all the units have a not null probability of being included in the sample.  
Called $s_{k,i}\in F_k, \forall k\in{\{1,\ldots,O\}}, \forall i \in{\{1,\ldots,n\}}$, with $n<<M$ a proper subset of the population data, $s_{k,i}\forall k,i$ is a \emph{sample} of the population and $\hat{\theta}(s_{k,i}) \forall k,i$ is an estimate of the true value $\Theta(f_{k,j})$ based only on the sample data.
How to choose the optimal sample is matter of research and the silver bullet has been the idea of sample representativeness: the sample should mimic as closely as possible the population characteristics, so to give reliable and precise estimates of the unknown true value of the statistics. 
If the sampling schema is grounded on probability theory (as most of the traditional methods actually are), then all units should have a not null probability of being included into the sample and probabilistic confidence intervals with controllable precision and confidence can be derived. This requires a very precise definition of the reference population and all the units to be reachable, Simple Random Sampling (SRS from now on) being the flagship technique \cite{wu2020sampling}. 

A probabilistic sampling strategy that controls precision and confidence of estimates is ideal, but in practice non probabilistic sampling is very common, due to the lack of enough data from the available population, to the cost of data gathering or simply to the easiness of scraping data from the web or buying/downloading pre-processed datasets ready to use.
When a non probabilistic sampling is chosen, the whole idea of representativeness of the sample is undermined. Samples can be built targeting costs, time, easiness or even results, purposely misleading estimators and classifiers towards confirmatory research.
This happens because the sampling process by itself introduces an uncontrollable \emph{selection bias}, consequence of the incomplete observation of the data, and because this bias can be manipulated once probabilistic design constraints are skipped.

In the following, \emph{sampling bias} will be considered as a special case of \emph{selection bias}, whereas the former is controllable, safe, and inherent to the probabilistic sampling design, while the latter is a generic bias due to the somewhat arbitrary choice of the units in the sample.

\subsubsection{Reservoir sampling}
One of the extensions of SRS to streaming data or infinite datasets, is the \textbf{reservoir sampling} \cite{ResSamp}. It solves the problem of selecting a random sample $[s_{k,i}]$ with finite size $n$, without replacement, from a pool of $N$ items where $M\geq N>>n$ is an unknown value (and so is the sampling fraction $n/M$).

The basic idea of reservoir sampling follows: 
\begin{enumerate}
    \item Save the first data candidates for the sample in a \emph{reservoir} of size $n$, one at time, until it is full. This is the temporary sample candidate $[s_{k,i}]$ with $i=j \ \forall j\leq n$.  

    \item For each subsequent instance $f_{k,j}, j>n$,  generate a random integer between 1 and $j$. Let this number be $\alpha$;
    
    \item If $\alpha$ is less than $n$, replace the  value in position $\alpha$ of the reservoir with the current datum, that is $s_{k,\alpha}=f_{k,j} \ \forall k$.
\end{enumerate}

All the data have the same probability of being included into the reservoir and the sample size is a free parameter. 
More efficient or weighted variations have been proposed lately \cite{ResL,ResW}. 

\subsubsection{Active Learning and sampling}
The recent paradigm of \emph{Active Learning} is grounded on the idea of allowing the learner to choose the data instances from which to learn (\cite{settles2009active,DeepActive}) and allows a significant reduction on the number of required labels.
Instead of using all the given data, or learning by the same amount from all the given data, the learner applies a selection criteria aimed to find the most informative instances and asks for labels to an oracle.

The process may become data-driven when an auxiliary or surrogate model is introduced and prediction on unseen data are used to  iterate between estimation and data collection with optimal
subsamples, the so called  \emph{active sampling} \cite{imberg2022active}.
It has been proposed as a way to improve the fairness of a learning algorithm by sampling more data from the worst-off classes \cite{abernethy2020active}, in order to improve the data collection process. 

\subsubsection{Oversampling minority class}
When dealing with imbalanced data, that is when one or more of the group or classes of data dominate the others in terms of volume, estimates are biased toward the majority group/class and the crude accuracy cannot be trusted as a metric of performance \cite{Mar1}.

One of the popular fix to the imbalance is at the data level, that is re-balancing the data undersampling the majority class or oversampling the minority class, or both, so to present a balanced training set to the classifier.

One of the most successful 
preprocessing techniques is called Synthetic Minority Oversampling Technique (SMOTE) , that generates new data by linear interpolation of random instances of the minority class. It has seen hundredths of variations so far (please see \cite{smote}).

\section{Samplation}
While a probabilistic sampling design would fix most of the issue at the data level  and would produce technically unbiased estimates, with a controllable precision and confidence, it would require a large effort in terms of time and money to gather data and cannot even be attempted for very large models or indefinite reference populations.

By consequence, most of the classifiers proposed and tested nowadays in the scientific literature are trained from standard "benchmark" datasets, affected by selection biases from the beginning. 
Free datasets from non probabilistic samples are readily available and continuously growing as benchmarks in repositories that are regularly used in scientific research.
Even when the training data are gathered \emph{ex novo}, the quickest and cheapest source of fresh data is through web scraping or voluntary participation, that unavoidably produces selection bias.

Moreover, Large Language Models are the trailblazer of models too heavy and big to be trained with commodity hardware: state-of-the-art Deep Neural Networks have billions of parameters and their large scale proprietary training requires GW of energy, so they can only be downloaded already trained to be fine-tuned before use.

Given this scenario, \emph{Samplation} is proposed here as a portmanteau word between "SAMPLing" and "data generATION" and refers to a technique that borrows ideas from oversampling minority class and active/reservoir sampling, with the aim of improving the fairness of a pre-trained classifier during fine-tuning.

As clarified above, a model trained with data affected by selection bias will be 
inevitably biased, it will have unreliable confidence intervals, and a probabilistic fix of a non probabilistic sampling design is unfeasible. 
On the contrary, samplation is non-probabilistic by design, it assumes selection bias: an artificial compensation bias is purposely introduced in the  data during fine-tuning of a pre-trained model to improve its fairness towards a target value. It should not be used if the training data come from a probabilistic sampling design.

\subsection{Considerations}
To be applicable, Samplation requires the following conditions to be fulfilled:
\begin{enumerate}
    \item the available training data should \emph{not} be a representative sample of the reference population, that is they should \emph{not} come from a probabilistic sampling design schema, in other words they should be affected by selection bias;
    \item the pre-trained model should have been trained on data that do \emph{not} come from a probabilistic sampling design, that is it should produce technically biased predictions;
    \item the trained model should give unfair predictions according to a given measure of fairness over one or more unprivileged groups; the desired level of fairness according to the chosen measure should be known \emph{a priori};
    \item the prediction accuracy should be considered secondary compared to the desired level of fairness, that is a technically biased estimator should be acceptable;
    \item the minority group/instances should be in the same order of magnitude of the majority instances, that is the maximum imbalance ratio among groups is greater than 1/10;
    \item at least a few examples exists in the minority group to be oversampled, that is artificial data cannot be manifactured from scratch over classes/groups that are not represented in the data.
\end{enumerate}

\subsection{Method}
Given a pre-trained model, a privileged group and an unprivileged group according to a discriminant variable, 
the key idea is the creation of \emph{reserves}, one for each possible value\footnote{Values can be binned or grouped if the number of distinct values is too high.} of the discriminant variable (i.e., the variable reflecting a bias). These value must be generated starting from the real instances with a data augmentation method, i.e. SMOTE.
Then an artificial sample of size $\tau<<N$ should be built sampling randomly from the reserves but with a reverse bias, to be used for fine tuning of the pre--trained model.

It can be considered both a \emph{pre} than a \emph{post} processing method, depending on the point of view: it ranks among the \emph{pre-processing} mitigation methods considering that it operates directly at the data level through a mixture of techniques involving \emph{data augmentation}; but it ranks among the \emph{post-processing} methods considering  that it operates during fine-tuning, after the model has been trained the first time.

\section{Experiments}
In the following, a concrete example of Samplation is  applied to the gender bias in a visual semantic role labeling problem. The creation of reserves is based on the generation of augmented images from the training set through geometrical transformations.
The imbalance ratio has been chosen to measure bias and the target value for fairness has been set to 1, that is the two groups should be perfectly balanced.

\subsection{Visual Semantic Role Labeling}
Visual semantic role labeling (vSRL) goes beyond the recognition of activities and human-object interactions, which aims to produce a concise summary of the situation depicted in an image, identifying the roles of each entity within the scene related to the corresponding activity to be performed, in terms of what is happening (e.g., clipping, cooking, adjusting, or whatever \emph{activity}), who is doing the action (\emph{agent} role), who is subjected to the action (\emph{patient} role), the instrument used (\emph{tool} role), the location (\emph{place}) and a large variety of other roles. 

To achieve this, visual semantic role labeling relies upon the English verb lexicon \emph{FrameNet} which pairs every verb with a \textbf{frame}\footnote{frame is meant to be a performed action to which semantic roles are associated. It is different from frames in the context of image processing.} composed by a set of \emph{semantic roles} (e.g., agent, patient, place, tool, source, destination etc.) whose aim is to specify how and which entities participate in the activity described by the verb. In combination with FrameNet, it is also used the lexical database \emph{WordNet} to fill the specific entities or objects in a given situation, previously defined by FrameNet. 

\begin{figure}[ht]
    \begin{center}
        \includegraphics[scale=0.45]{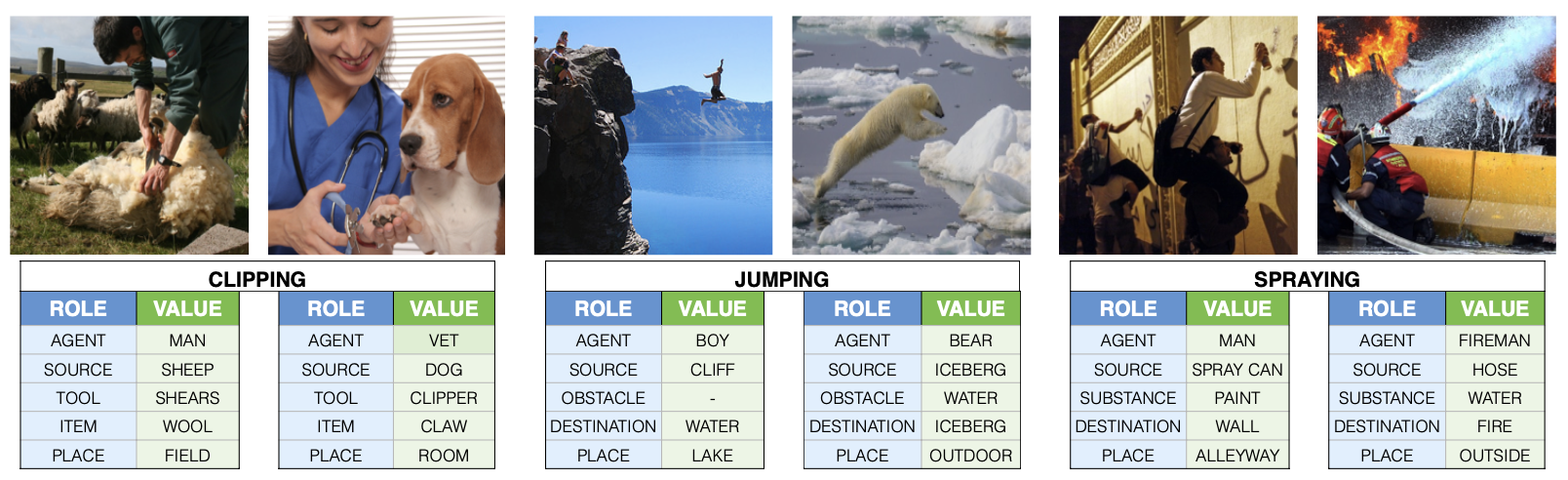}
        \caption{Example of Visual Semantic Role Labeling. Image source: \cite{yatskar2016situation}}
        \label{fig:example}
    \end{center}
\end{figure}

To formally define vSRL we assume a discrete set of verbs $V$, nouns $N$ and frames $F$. Specifically, each frame $f \in F$ is composed by its discrete set of semantic roles $S_f$ (e.g., agent, patient, source, tool, etc.). Semantic roles can be also shared across different frames. 
For instance, if we consider the \textcolor{blue}{Figure~\ref{fig:example}}, the semantic roles $S_f$ related to the first image are: \emph{agent}, \emph{source}, \emph{tool}, \emph{item} and \emph{place}. We can easily note that the \emph{agent} semantic role is shared among the three different frames, while some other semantic roles (e.g., \emph{item}, \emph{obstacle}, \emph{substance}, etc.) appear just in specific frames.

In turn each semantic role $s \in S_f$, related to a certain frame $f$, is fulfilled by a noun value $n_s \in N \cup \{\varnothing\}$ that could also assume no value. 

The pairs of semantic roles and their corresponding values (i.e., nouns) is denoted as \textbf{realized frame} $R_f = \{(s, n_s) : s \in S_f\}$. Looking at \textcolor{blue}{Figure~\ref{fig:example}}, the realized frame for the first image is: $R_f =$ \{(\emph{agent}, \emph{man}), (\emph{source}, \emph{sheep}), (\emph{tool}, \emph{shears}), (\emph{item}, \emph{wool}), (\emph{place}, \emph{field})\}. \\
Defined the previous notations, the aim of  vSRL is to predict a situation $T$ depicted in an image such that $T = (v, R_f)$, which is a pair involving the verb $v$ ($v\in V$) and its corresponding realized frame $R_f$ \cite{yatskar2016situation} i.e., to pair semantic roles and their associated nouns.

Given an image $i$, the most common solutions to predict the corresponding situation $T = (v, R_f)$ involve a \emph{Neural Conditional Random Field} as baseline model. Specifically, the study (conducted by Taori and Hashimoto \cite{taori2022data}) employs a \textbf{Conditional Random Field} (\emph{CRF}) model backed with \textbf{ResNet}. 
By means of the ResNet architecture, the hierarchical features are extracted from each image, which will be useful for capturing the \emph{verb-role-noun} triplets occurring in them. Hence these features, extracted by the deep neural network, predict factors in the CRF model. 

On the other hand, CRF is a probabilistic graphical model for labeling and segmenting \emph{structured data} based on an undirected graph $G = (V, E)$ where $V$ is the set of vertices, in terms of variables, and $E$ is the set of edges connecting these nodes. CRFs are an extension of \emph{HMMs} in which some constraints on transition probabilities and conditional dependence are relaxed. This model falls under \emph{discriminative models} which aim to estimate conditional probabilities between labels and input data, without arriving at a complete knowledge of the underlying data generation processes. 

CRF is trained according to the \textbf{maximum likelihood} estimation, so it seeks to find the model parameters $\theta$ that maximize the \emph{conditional probability} $P(y|x)$ of the output variables given the input variables, where the output variables correspond to the \emph{label sequences} and the input variables are the \textbf{image data}, for this specific context. 

\subsection{Dataset}
To approach the vSRL task, a benchmark dataset called \textbf{imSitu}\footnote{https://prior.allenai.org/projects/imsitu} has been chosen. It offers a rich collection of over 125.000 images depicting 200.000 distinct situations. Each situation involves one of 504 possible actions (or \emph{verbs}) and values for up to 6 activity-specific roles. 
Specifically, there are 1.788 unique semantic roles with 190 distinct types based on their similarities (i.e., the number of distinct classes in which similar roles can be grouped). For each image, the imSitu dataset contains 3 different situations, of which 205.095 are unique situations.
The images were retrieved from Google Image search with query expansion techniques and labeled with detailed information about the actions being performed by individuals or objects within the scenes on Amazon Mechanical Turk.
Furthermore, the Dataset is already split in training, validation (or development) and test set for serving each specific stage of Machine Learning.

\vspace{1em} \begin{table}[ht]
\begin{center}
\begin{tabularx}{0.94\textwidth} { 
  | >{\raggedright\arraybackslash}X 
  | >{\centering\arraybackslash}X 
  | >{\raggedleft\arraybackslash}X | }
\hline
\textbf{verbs} & 504 \\
\hline
\textbf{images} & 126.102 \\
\hline
\textbf{situations per image} & 3 \\
\hline
\textbf{total annotations} & 1.481.851 \\
\hline
\textbf{unique roles (role types)} & 1788 (190) \\
\hline
\textbf{images per verb (range)} & 250.2 (200 - 400) \\
\hline
\textbf{unique realized frames ($\geq$ 3)} & 205.095 (21.505) \\
\hline
\textbf{train / dev / test split} & 75.702 / 25.200 / 25.200 \\
\hline
\end{tabularx}
\caption{Summary statistics of imSitu dataset.}
\end{center}
\end{table}

\subsection{gender bias}
The ImSitu dataset is known for exhibiting a gender bias on some verbs (for example "cooking"). 
To obtain a controlled experiment, the data to train the first time the classifier are  sampled with a prevalence of the privileged class, with an increasing percentage (60\%-40\%; 70\%-30\%;80\%-20\%;90\%-10\%). 
The data for subsequent fine-tuning are then sampled from each of the two reserves in inverse percentages according to the \textbf{reservoir sampling} strategy: sampling from the female reservoir will be done considering the male prediction percentage and vice versa. The sample size for fine-tuning  will range in the 400-1000 range, providing for denser intervals between 700 and 800. 
First the classifier is pre-trained with 20k training data, then a second round of training (fine tuning) is performed using the reversely-biased samples re-evaluating the percentage of male and female predictions in the test set after this step of fine-tuning.

\begin{figure}
    \begin{center}
        \includegraphics[scale=0.31]{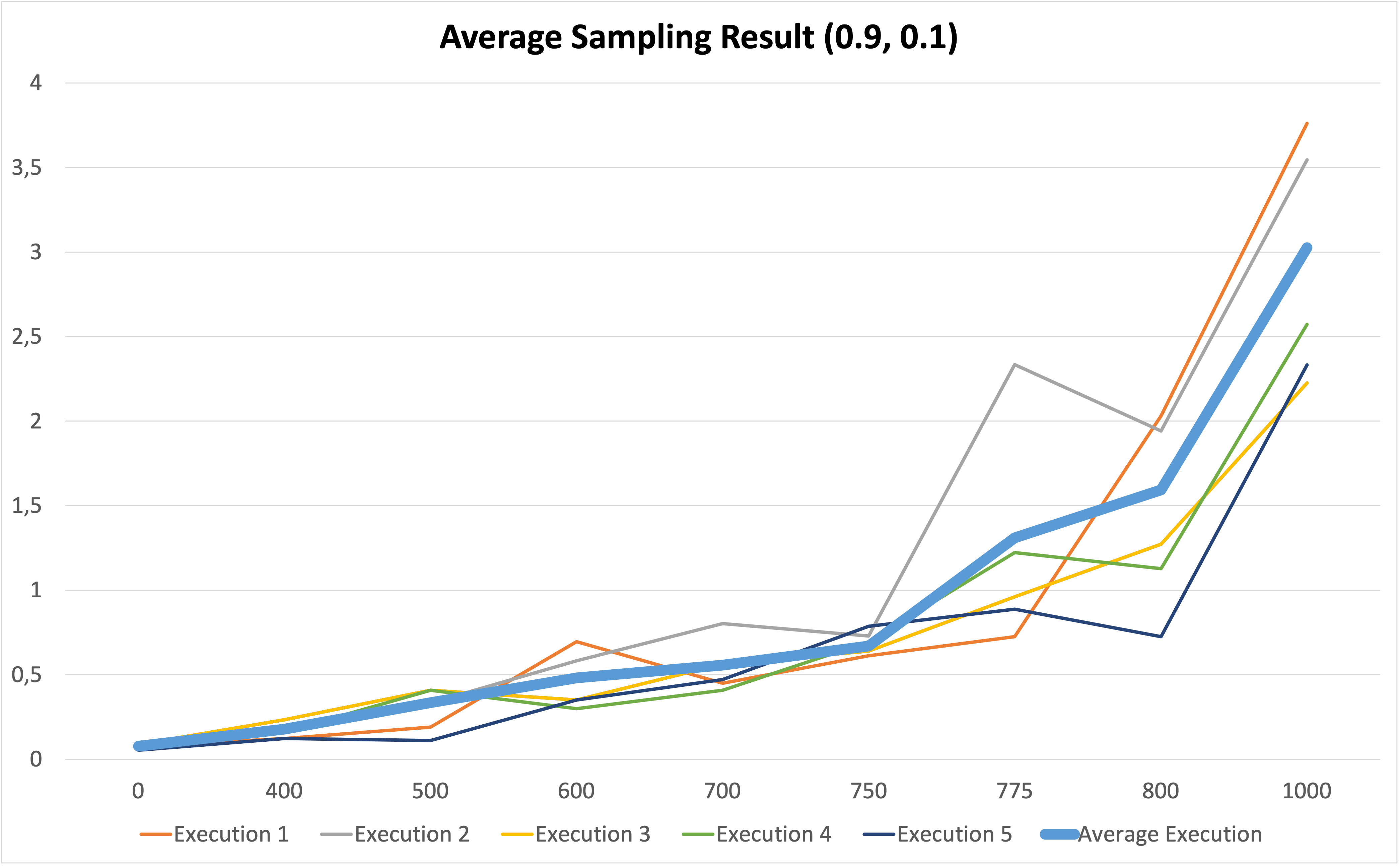}
        \caption{Test-time results on the model predictions of the proposed method when the initial imbalance is 90\%-10\%. $X$ axis, sample size; $Y$ axis, imbalance ratio. Bold line represents the average.}
        \label{fig:result1}
    \end{center}
\end{figure}

\begin{figure}
    \begin{center}
        \includegraphics[scale=0.31]{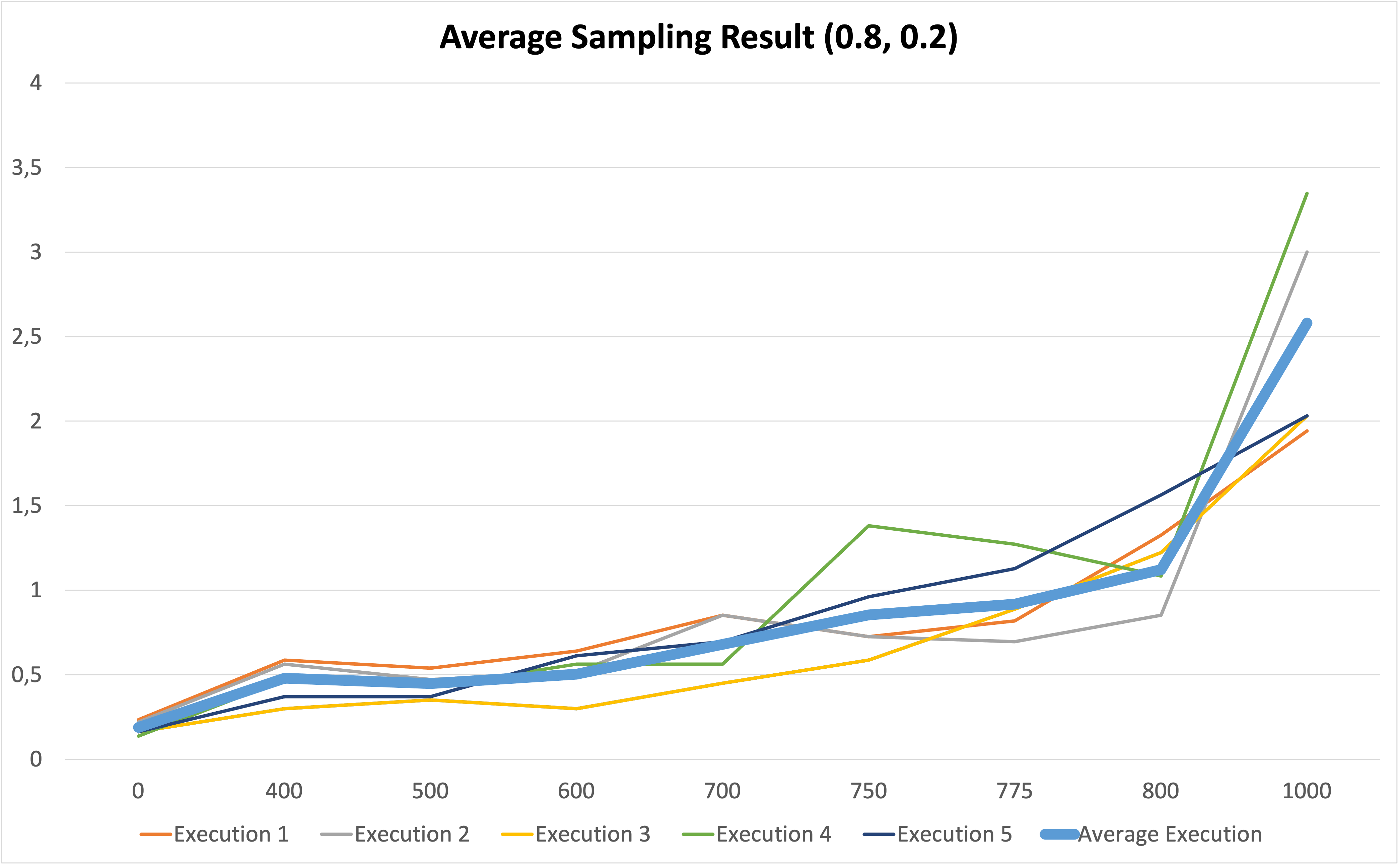}
        \caption{Test-time results on the model predictions of the proposed method when the initial imbalance is 80\%-20\%. $X$ axis, sample size; $Y$ axis, imbalance ratio. Bold line represents the average.}
        \label{fig:result2}
    \end{center}
\end{figure}

\begin{figure}[ht]
    \begin{center}
        \includegraphics[scale=0.31]{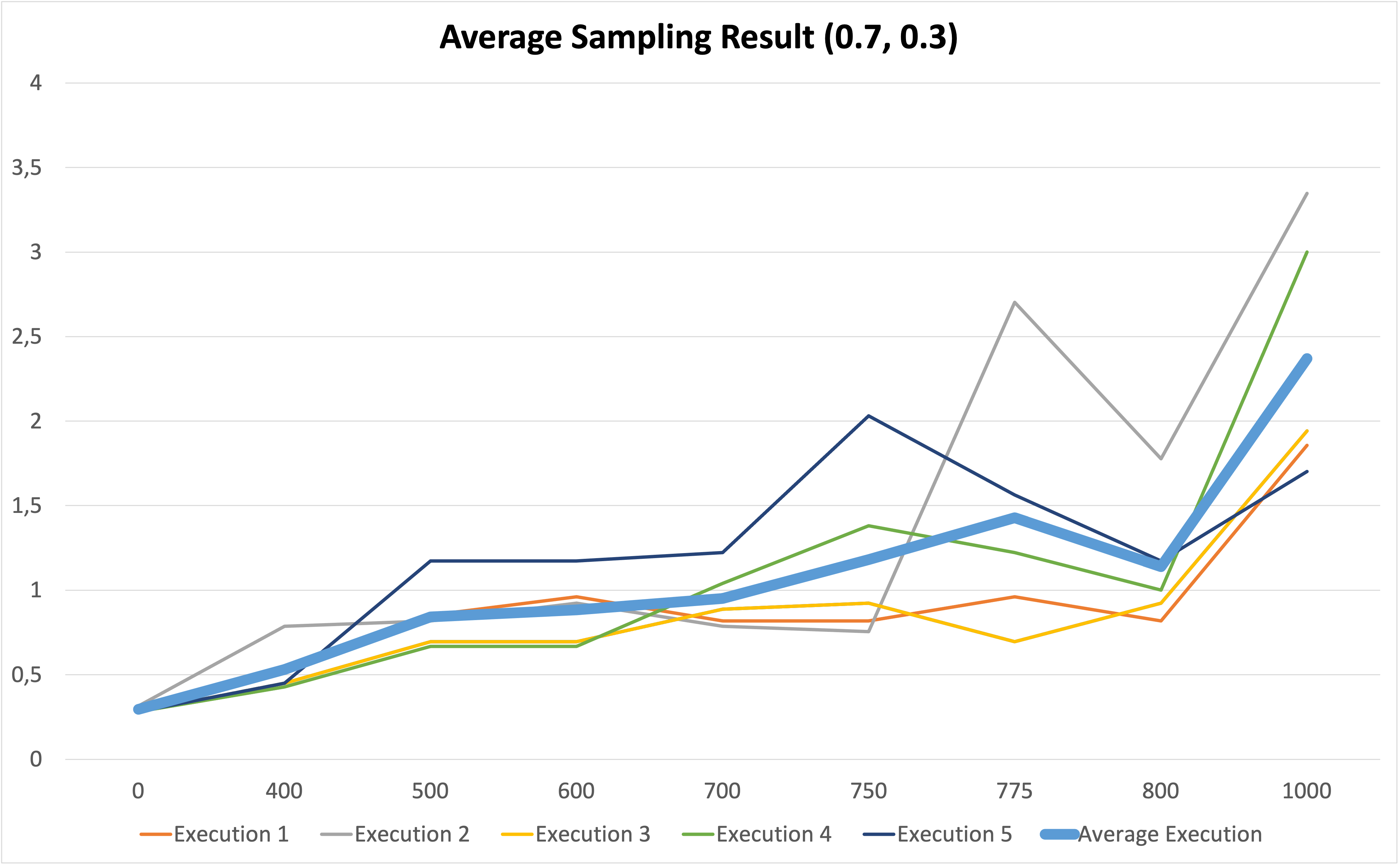}
        \caption{Test-time results on the model predictions of the proposed method when the initial imbalance is 70\%-30\%. $X$ axis, sample size; $Y$ axis, imbalance ratio. Bold line represents the average.}
        \label{fig:result3}
    \end{center}
\end{figure}

Results can be seen in figures \ref{fig:result1}, \ref{fig:result2} and \ref{fig:result3}. 
Perfect balance happens when the ratio ($Y$-axis) is around 1.
Reserves obtain the \textbf{annotations} of the artificially generated images copying  the annotations of the corresponding original images contained in the dataset, and their size is around 16,000. 

The additional training data should not over-correct, as the optimal correction is achieved when the ratio between the two percentages is around the target value (1 in this case, that represents perfect balance) and, once reached, the training should stop otherwise there is the risk of inducing a reverse bias, that is a totally reversed proportion of predictions, by which the privileged class becomes the previously unprivileged one.

The ideal size of the reversely biased sample has been only empirically estimated and it depends on the severity of the initial bias to be cured:
in the performed tests, perfect fairness happens with a sample size $\tau$ between 500 and 750, with less than 3\% of worsening of the mean accuracy (as expected due to the well known fairness-accuracy trade-off).

\begin{figure}[ht]
    \begin{center}
        \includegraphics[scale=0.27]{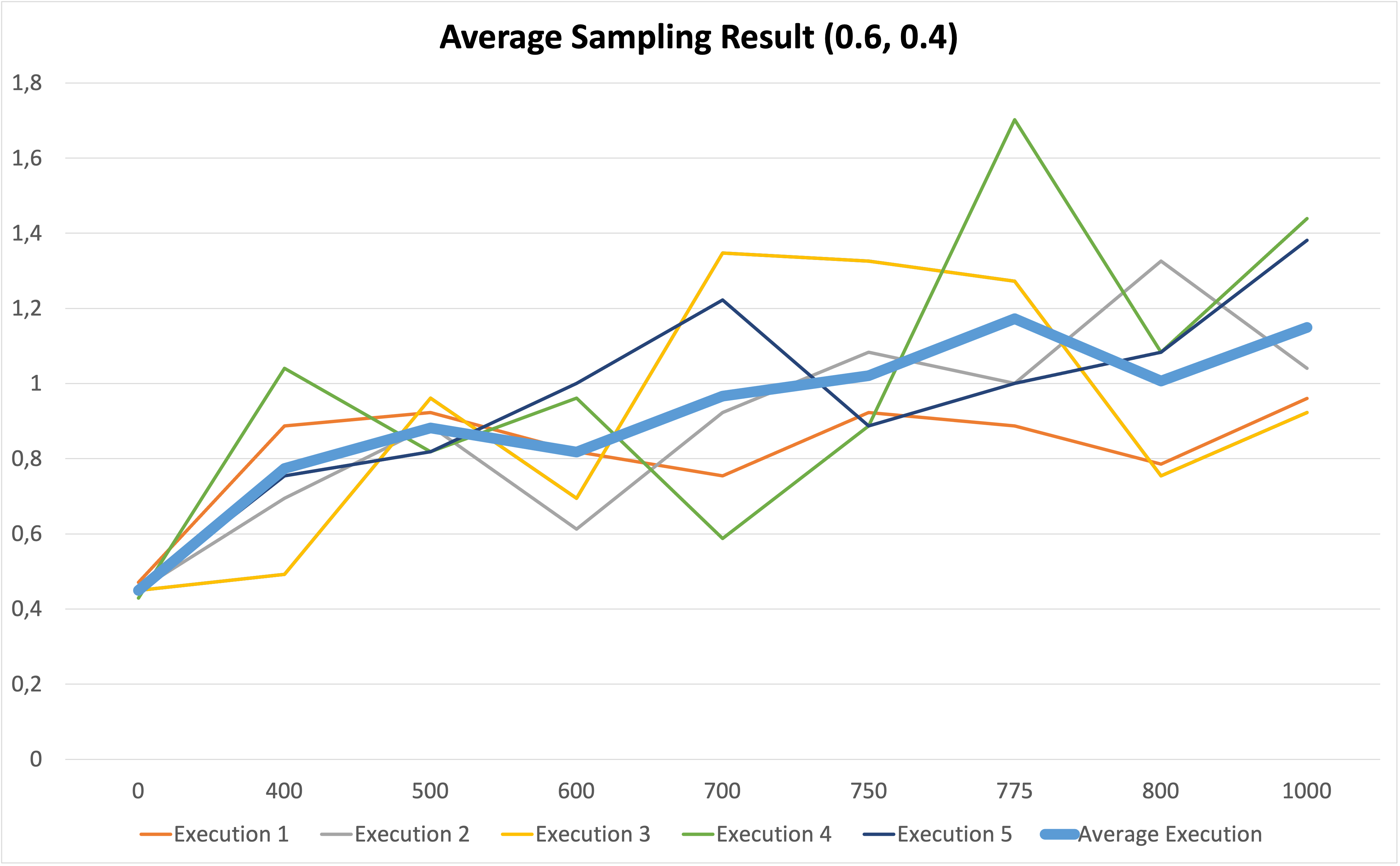}
        \caption{Test-time results on the model predictions of the proposed method when the initial imbalance is 60\%-40\%. $X$ axis, sample size; $Y$ axis, imbalance ratio. Bold line represents the average.}
        \label{fig:result4}
    \end{center}
\end{figure} 

As can be seen from figures and as expected, samplation is less effective when the predictions before fine-tuning are already almost balanced (60\%-40\%, see figure \ref{fig:result4}). On the other side, the more the predictions before fine-tuning are imbalanced, the more effective it is is (see figures \ref{fig:result1},\ref{fig:result2},\ref{fig:result3}). It is worth noting that very few samples are sufficient to remediate a moderate to a severe bias (500-700 instances compared to 20.000 in the original training); that the optimal value of sample size (the one that produces perfect balance) is quite stable and grows less than proportionally increasing the imbalance ratio; that it is critical not to exceed the optimal value of sample size because of the abrupt bias reversing that happens, over-correction. The experiments suggest a conservative choice of this size.

\section{Conclusions}
A new pre-processing method to full remedy unfair classification outcomes, called \emph{samplation}, has been proposed and tested on a visual semantic role labeling problem.
The method generates different reserves, based on the possible values or subgroups of the discriminant variable, and builds new training samples from these reserves with a reversed bias.
Using as test case the gender bias in visual semantic role labeling, it has been shown experimentally that a small percentage, less than 3\%, of artificial data selected at inverse percentages from reserves during fine-tuning is sufficient to fully cure an initial imbalance of up to 9 to 1 on the prediction of the unprivileged group, with a moderate effect on accuracy. 
Further study are required to confirm these findings, to extend them to contrasting bias amplification in reinforcement loops and to estimate the necessary sample size that avoids over-correction.

\bibliographystyle{splncs04}
\bibliography{mybibliography}

\end{document}